\documentclass{acm_proc_article-sp}
\begin{document}


\title{Retrieval from Captioned Image Databases Using Natural Language Processing}
\numberofauthors{1}
\author{
\alignauthor David Elworthy\titlenote{Now at: Microsoft Research Limited, St
George House, 1 Guildhall Street, Cambridge CB2 3NH, United Kingdom}\\
  \affaddr{Canon Research Centre Europe, Guildford, United Kingdom}\\
  \email{dahe@acm.org}
}

\maketitle
\begin{abstract}
At first sight, it might appear that natural language processing should
improve the accuracy of information retrieval systems, by making available a
more detailed analysis of queries and documents. Although past results appear
to show that this is not so, if the focus is shifted to short phrases rather
than full documents, the situation becomes somewhat different. The ANVIL
system uses a natural language technique to obtain high accuracy retrieval of
images which have been annotated with a descriptive textual caption. The
natural language techniques also allow additional contextual information to be
derived from the relation between the query and the caption, which can help
users to understand the overall collection of retrieval results. The
techniques have been successfully used in a information retrieval system which
forms both a testbed for research and the basis of a commercial system.
\end{abstract}


\newcommand{\aum}{\"{a}}

\section{Introduction}

Text information retrieval is concerned with finding documents which match
against user's query, and assigning a measure according to the closeness of
the match. Natural language processing (NLP) can provide rich information
about the text, and it might appear reasonable that this would result in
better retrieval than conventional ``bag of words''
approaches. Fagan \cite{Fagan:1987} reports experiments in which simple
keywords were augmented with compound terms consisting of pairs of
keywords. While the addition of compound terms produced better accuracy, no
significant difference was observed between terms selected on the basis of
their linguistic relationship and ones selected purely on the basis of their
statistical association. Smeaton \cite{Smeaton:1997} makes similar observations and
on the basis of a number of experiments concludes that NLP has little to offer
IR. However, there are exceptions in some niche areas. For example,
Flank \cite{Flank:1998} describes a retrieval system in which the ``documents''
are short image captions. She uses the techniques of searching on heads and
head-modifier combinations introduced by Strzalkowski \cite{Strz:1994}, and obtains high
precision and recall. It therefore appears that in specialised applications, NLP
may have something to offer.

Here we will look at a technique called {\em phrase matching}, which attempts
to use lightweight, symbolic natural language analysis to improve retrieval
accuracy. Like Strzalkowski's work, it relies on looking for combinations of
words which stand in certain modification relationships, and like Flank, we
have applied it to searching a database of annotated images. It differs from
the earlier work in two important ways. Firstly, it does not simply use the
analysis of the captions and queries as a source of compound terms, as
Strzalkowski does. Instead it
recursively explores the structure of the caption and query, checking that
terms stand in equivalent modification relations in the two phrases. This also
allows the match score to be finely tuned and special cases such as negation
to be handled. Secondly, by means of a further algorithm called {\em context
extraction}, information about non-matching parts of the caption, related
to the parts which did match, can be obtained. The retrieval results can then
be organised and categorised by the contexts they have in common, with the
goal of helping users of the retrieval system to understand and organise the
results. This is an important step, because it provides information which is
unavailable without natural language analysis, and shows that NLP can
contribute in adding new functionality as well as improving accuracy.

In section~2 of this paper, we will introduce the phrase matching algorithm,
and give some evaluation results. Section~3 then moves on to context
extraction. Some conclusions and suggestions for future work are presented in
section~4. We first briefly look at the application for which the work was
intended.

\subsection{The ANVIL system}
ANVIL (Accurate Natural Language Visual Information Locator) is a retrieval
system for databases of digital photographs, intended for operation over the
world wide web. The photographs are annotated with captions, typically between
10 and 30 words in length, which describe the subject matter of the image. The
system is intended for casual users, and it is therefore important to make it
easy to formulate and refine queries, and to help the users understand the
results. This is the main motivation for using phrasal captions and phrase
matching: while traditional IR techniques over collections of keywords may
give good recall, they are not really suitable for users who will give up if
they do not see an acceptable result in the first few presented by the
system. ANVIL is further enhanced by an interactive user interface, details of
which can be found in Rose et al. \cite{Rose:2000}.

In outline, the processing in ANVIL proceeds as follows. When images are
registered with the system, their captions are analysed into a meaning
representation. The terms from the captions are stored in an index database,
pointing to records containing the image identifier and the analysed
caption. In retrieval, the terms are extracted from the query and used to find
candidate captions using conventional IR techniques such as vector-cosine
matching or the similarity techniques of Smeaton and Quigley \cite{SQ:1996}; this phase is
called simple matching. The query is analysed to a meaning representation in
the same way as the captions, and the representations of the query and
candidate captions are compared using natural language matching techniques. The
result of the comparison is a score, which is combined with the score from
simple matching. Contexts may also extracted at this stage, and the resulting
images with their scores, captions and contexts are presented to the user.

\section{Phrase matching}

The basic idea in phrase matching is as follows. We start by analysing the
query and the caption into dependency structures, in which the words are
connected by labelled links indicating the relationship between them. One word
(or occasionally more) will not be a modifier of any other words. It is
designated the head, and is the word which says, in most general terms, what
the caption is about. The head of the query is compared against words in the
caption, starting from its own head and progressing to modifiers if no match
is found. If there is a match, the modifiers of the query head are compared
against modifiers of the corresponding term in the caption. For each word that
matches, the process recurses in a similar way down through the dependency
structure. The modification relationships can be simple ones, or they can
involve tracing through several dependency links. Each stage of the comparison
has a score associated with it, so that strong and weak matches can be
assigned different scores. Finally, we allow matching of elements in the
dependency structure against fixed expressions, to detect special cases such
as negation.

\begin{figure} 
\centering
\setlength{\unitlength}{1cm}
\begin{picture}(12,6.5)(0,1)
\put(0,1.5){\makebox{\rm colour}}
\put(2,1.5){\makebox{\rm document}}
\put(3.9,1.5){\makebox{\rm copier}}
\put(1.6,2){\oval(1.7,1)[t]}
\put(1.5,2.5){\vector(1,0){0.3}}
\put(1.3,2.7){\makebox{\tt mod}}
\put(3.3,2){\oval(1.7,1)[t]}
\put(3.2,2.5){\vector(1,0){0.3}}
\put(3.0,2.7){\makebox{\tt mod}}
\put(0,4.5){\makebox{\rm copier}}
\put(2.0,4.5){\makebox{\rm for}}
\put(3.9,4.5){\makebox{\rm colour}}
\put(5.8,4.5){\makebox{\rm documents}}
\put(1.2,5){\oval(2,1)[t]}
\put(1.3,5.5){\vector(-1,0){0.3}}
\put(0.8,5.8){\makebox{\tt prep}}
\put(4.2,5){\oval(4,2)[t]}
\put(4.2,6){\vector(-1,0){0.3}}
\put(3.2,6.2){\makebox{\tt phead}}
\put(5.2,5){\oval(2,1)[t]}
\put(5.2,5.5){\vector(1,0){0.3}}
\put(4.8,5.7){\makebox{\tt mod}}
\end{picture}
\caption{Example dependency structures}
\label{f1}
\end{figure}
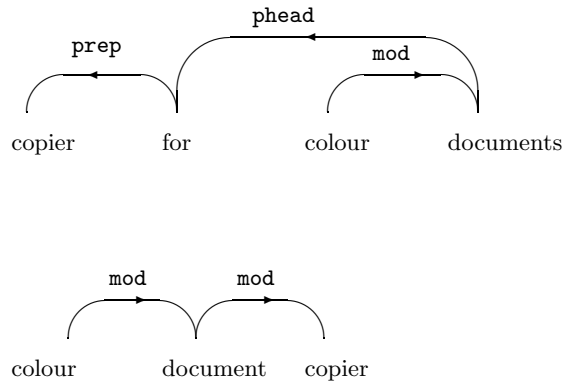

Figure~\ref{f1} shows the dependency structures for two phrases with similar
meanings.
Dependencies are shown as pointing from a modifier to the term it
modifies. Although dependency structures go some way to abstracting away from
the syntactic analysis, we still need a way of assigning a similarity between
non-identical structures. In this example, we want the noun-noun modification
between {\em copier} and {\em document} in the lower phrase to have a high
similarity to the modification via the preposition {\em for} in the upper one.

For convenience, we represent dependency structures using a notation of
indexed variables, in which the name of the variable stands for the name of the
dependency, and the variable is indexed on the modified word. An unindexed
variable is used for the head. The examples can then be written as
\begin{verbatim}
colour document copier
    head = copier
    mod[copier]   = document
    mod[document] = colour

copier for colour documents
    head = copier
    prep[copier]   = for
    phead[for]     = documents
    mod[documents] = colour
\end{verbatim}
Thus, for example, \texttt{mod[copier] = document} indicates that {\em copier}
stands in the \texttt{mod} relation to {\em document}, i.e the \texttt{mod}(ifier)
of {\em copier} is {\em document}.

Dependency structures are especially suitable for this kind of
processing. They are closely related to the syntactic form, but abstract away
from the linear order of the words and fine details of phrase structure. From
a practical point of 
view, dependency structures can be computed quickly and efficiently; see for
example, the dependency parser built by J\"{a}rvinen and Tapanainen \cite{Jarvinen:1997} or the Link
grammar parser of Sleator et al. \cite{Sleator:1991}. We use a finite-state parser which
has been modified to deliver the dependencies as well as the phrase bracketing
(Elworthy \cite{Elworthy:2000}). It works in time roughly proportional to the square
of the number of words in the phrase.

\subsection{Matching rules}

A system of rules specifies what relationships can be treated as equivalent. A
small set of example rules appears in figure~\ref{f2}.
\begin{figure} 
\centering
\begin{verbatim}
head_rule
{
 head = head  1.0 => mod_rule 0.7;
 head = mod[] 0.5 => mod_rule 0.7;
 mod[]  ?     0.3 => Done 1.0;
}

mod_rule
{
 mod[] = mod[] 1.0 => mod_rule 1.0;

 phead:prep[] = phead:prep[] 1.0 => mod_rule 1.0;
 phead:prep[] = mod[] 1.0 => mod_rule 1.0;
 mod[] = phead:prep[] 1.0 => mod_rule 1.0;

 vhead:cop:rel[]
     = vhead:cop:rel[] 1.0 => mod_rule 1.0;
 vhead:cop:rel[] = mod[] 1.0 => mod_rule 1.0;
 mod[] = vhead:cop:rel[] 1.0 => mod_rule 1.0;

 amod[] = amod[]  1.0 => Done 1.0;
 'not' = amod[]  0.0 => Done 0.0;
 amod[] = 'not' 0.0 => Done 0.0;
}
\end{verbatim}
\caption{Example matching rules}
\label{f2}
\end{figure}
The left and right hand sides of a comparison express paths through the
dependency structure. The idea is that if we have already found a query
word which matches a word from the caption, we then follow the specified paths
from these words, and compare the words lying at the end of the paths.

It is convenient to gather rules into named groups, such as \texttt{head\_rule}
and \texttt{mod\_rule}. One group is designated the start group, and its rules
are applied to start the matching process. Within a group, the rules are
applied in order, so that later rules in a group can be used to test words
which were not caught by the earlier rules. Each rule has a continuation,
which specifies what should happen after it has been applied. As an example,
consider the rule
\begin{verbatim}
head = head    1.0 => mod_rule 0.7;
\end{verbatim}
This says that after matching head words, continue with the rule
group \texttt{mod\_rule}. The words which have just matched provide the
starting point for paths in the continuation; in effect they are substituted
where \texttt{[]} appears in \texttt{mod\_rule}. The special continutation {\tt
Done} indicates that no further comparison is to be carried out from the words
that matched.

The process is started by comparing words without indexing, stored in {\tt
head}. Thus, the structures in the examples of figure~\ref{f1} can be matched
by starting with \texttt{head = head} and then continuing with \texttt{mod[] = phead:prep[]},
indicating that a modifier of the head (\texttt{mod[]}) can be compared with the
head of a prepositional phrase (\texttt{phead}) reached by following from a
matched caption word a preposition (\texttt{prep[]}).

There are two special sorts of rules: mopping-up rules and token
rules. Mopping-up rules specify that certain words are to be considered to
have matched, without actually consuming any words from the other phrase. One
use is to catch words from the query which did not have a counterpart in the
caption. For example,
\begin{verbatim}
mod[] ?     0.3 => Done 1.0;
\end{verbatim}
causes modifiers from the query to be mopped up\footnote{Since this is
in the start rule group, the whole unmatched range of the \texttt{mod}
variable is used, without indexing.}. Token rules allow matching against
specific words. For example, the rule
\begin{verbatim}
'not' = amod[]  0.0 => Done 0.0;
\end{verbatim}
allows an \texttt{amod} (``adverbial'' modifier) in the caption to be tested
against the literal word {\em not}, with an effect on scoring described below. There are a few
further variants of rules which we will not discuss here, for example rules
with a negated test, and ones which are sensitive to word order.

\subsection{The scoring scheme}

The scoring scheme is a critical part of phrase matching, as it will allow us
to distinguish exact and near-exact matches from partial and weak ones. The general
approach is to assign each word of the query phrase two numeric values, called
the {\em score} and the {\em weight}. The score of a query word is a measure
of how well it matched considered in isolation from the rest of the caption,
while the weight indicates the importance of the rule application. In general,
words which are compared in the start rule group, such as the head, will be
more important than ones compared as a result of a continuation, such as
modifiers. Scores are assigned to query words if they actually matched a word
in the caption, or if they were caught by a mopping up rule or token rule. The
score does not take the caption words into account, other than an allowance
for their similarity with query words. A special score, called an {\em
up-score} is also used to handle words at the end of paths for special cases
such as negation. Writing the scores as $s_{i}$ and the weights as $w_{i}$,
the overall score of the match is $\sum s_{i}w_{i} /\sum s_{i}$, modified by
the up-scores as described below.

The rules are annotated with two values, called the $t$ (term) factor and the
$d$ (down) factor. In general, the $t$-factor provides the basic score for
words which were matched by the rule, and the $d$-factor sets the weight for
continuations. Thus, in
\begin{verbatim}
head = mod[]   0.5 => mod_rule 0.7;
\end{verbatim}
the $t$-factor is 0.5 and the $d$-factor is 0.7.

At the start of matching, the weight is 1.0. As we follow through
continuations, it is the product of the $d$-factors of the rules leading to
this point. If the rule above were in the start group, the weight of words
matched in \texttt{mod\_rule} would be 0.7, and if \texttt{mod\_rule} contained
\begin{verbatim}
mod[] = mod[]   1.0 => submod_rule 0.6;
\end{verbatim}
then the weight in \texttt{submod\_rule} would be $0.7\times 0.6$. The scores are
formed from the product of the $t$-factor of the rule, and two special
factors. Firstly, the similarity between the words can be used. For example,
we might allow {\em car} to match {\em vehicle}, but with a reduced
score. This factor could be calculated using lexical similarity metrics such
as those of Resnik \cite{Resnik:1995} or Jiang and Conrath
\cite{JiCon:1997}. A further extension
would be to recognise that the agentive suffix {\em X-er} (as in {\em copier})
allows a
match against the whole phrase {\em machine for X-ing} (as in {\em machine for
copying}), and similar rules based on derivational morphology. We do not take
this step in the current version of phrase matching.

The second special factor is the up-score. When a \texttt{Done} continuation
is reached,
its $d$-factor is multiplied into the score assigned by the rule which invoked
it\footnote{This represents a harmless overloading of the rule
notation}. Usually the factor will be 1.0, but in special cases it may be some
other 
value. An example of where this is useful can be found in the rules involving
`not' in figure~\ref{f2}. When a negation is seen, we effectively cancel the
score on the word which is negated, by using a $d$-factor, and hence an
up-score, of 0. Note that
making this kind of adjustment based on word pairs without the recursion
through the overall structure, as in Fagan's and Strzalkowski's work, is
very hard to do.

To show the rules in operation, suppose the query {\em yellow car} is tested
against {\em yellow car}, {\em car which is yellow} and {\em car which is not
yellow}. The dependency structures, written as variables, are shown in
figure~\ref{f3}, and a trace through the matching process appears in
figure~\ref{f4}.
\begin{figure} 
\centering
\begin{verbatim}
yellow car
     head = car
     mod[car] = yellow

car which is yellow
    head = car
    rel[car]   = which
    cop[which] = is
    vhead[is]  = yellow

car which is not yellow
    head = car
    rel[car]     = which
    cop[which]   = is
    vhead[is]    = yellow
    amod[yellow] = not
\end{verbatim}
\caption{Dependency structures for the matching example}
\label{f3}
\end{figure}
In particular, note how the rule
\begin{verbatim}
'not' = amod[]  0.0 => Done 0.0;
\end{verbatim}
causes the previous score assigment for {\em yellow} to be replaced by 0 when
comparing against {\em car which is not yellow}.
\begin{figure*} 
\centering
\textbf{yellow car + yellow car}\\
\begin{tabular}{|l|l|l|c|c|} \hline
Query word & Rule group & Comparison & Score & Weight \\ \hline
car    & head\_rule & head = head   & 1.0 & 1.0 \\
yellow & mod\_rule  & mod[] = mod[] & 1.0 & 0.7 \\ \hline
\end{tabular}
\\Overall match score = $(1.0\times 1.0 + 1.0\times 0.7) / (1.0 + 0.7) = 1.0$
\vspace*{0.5cm}\\
\textbf{yellow car + car which is yellow}\\
\begin{tabular}{|l|l|l|c|c|} \hline
Query word & Rule group & Comparison & Score & Weight \\ \hline
car    & head\_rule & head = head             & 1.0 & 1.0 \\
yellow & mod\_rule  & mod[] = vhead:cop:rel[] & 1.0 & 0.7 \\ \hline
\end{tabular}
\\Overall match score = $(1.0\times 1.0 + 1.0\times 0.7) / (1.0 + 0.7) = 1.0$
\vspace*{0.5cm}\\
\textbf{yellow car + car which is not yellow}\\
\begin{tabular}{|l|l|l|c|c|} \hline
Query word & Rule group & Comparison & Score & Weight \\ \hline
car    & head\_rule & head = head             & 1.0 & 1.0 \\
yellow & mod\_rule  & mod[] = vhead:cop:rel[] & 1.0 (initially) & 0.7 \\
(none) & mod\_rule  & 'not' = amod[]          & 0.0 & 0.0 \\
yellow & mod\_rule  & mod = vhead:cop:rel[]   & 0.0 (on up-score) & 0.7 \\ \hline
\end{tabular}
\\Overall match score = $(1.0\times 1.0 + 0.0\times 0.7) / (1.0 + 0.7) = 0.59$
\caption{Matching in action}
\label{f4}
\end{figure*}
The scores in this rule set are chosen on the basis of examining a variety of
examples, some of which might be expected to provide a close match, some a
partial match, and some a weak match. No experiments on learning the scores
from data have been carried out.

\subsection{Evaluation}

Evaluation of image caption retrieval is limited by the lack of suitable large test collections. We therefore created our own captions for a
set of digital photographs. The captions were prepared according to a set of
guidelines, so that they emphasised the objects in the image rather than
layout or composition. The guidelines were formulated to overcome problems with quality which
had been seen both in a pilot study, and the captions used by
Smeaton and Quigley \cite{SQ:1996}. There were 1932 captions in the set, with lengths ranging
from 1 to 22 words (9.0 average). Almost all of the captions were noun
phrases. It is relatively easy to construct a grammar which correctly analyses
all the phrases.

A query set was constructed by taking pictures from another source, and
devising phrases which should elicit a related image. An initial set of
results was obtained by pooling several keyword-based retrieval runs,
discarding queries which produced no results\footnote{With such a small test
collection and using a single retrieval systems, it might have been better to
construct complete relevance judgements rather than use pooling. However, time
pressures obviated doing this.}. The top results from phrase
matching with each query were then judged for relevance by two human
assessors, acting separately. Neither assessor was responsible for writing the
captions; one of them devised the queries. A standard precision-recall measure
was then calculated, using the TREC interpolation procedure (from {\tt
http://trec.nist.gov/}). An example of the output for a query, showing some
sample captions appears in figure~\ref{fout}.

The main comparison point between different tests was chosen to be the
precision at 10\% recall. This represent the case of naive or casual users,
who do not care about completeness in the results and who want high accuracy
in the first few (Pollock and Hockey \cite{pollock}). The precision at 5
documents and the R-precision were also calculated, although they are less
useful, partly there is often a very small number of
relevant results in such a small test set. Table~\ref{t1} shows the results
for a simple weighted
keyword matching strategy, and for phrase matching, using the two sets of
relevance judgements.
\begin{table*} 
\centering
\begin{tabular}{|l|c|c|c|} \hline
Run (assessor) & Precision at & Precision at & R-precision \\
    & 10\% recall  & 5 documents  &             \\ \hline
Simple matching (I)  & 85\% & 45\% & 61\% \\
Phrase matching (I)  & 92\% & 46\% & 66\% \\
Simple matching (II) & 87\% & 49\% & 63\% \\
Phrase matching (II) & 95\% & 53\% & 72\% \\ \hline
\end{tabular}
\caption{Evaluation results}
\label{t1}
\end{table*}

Phrase matching produces a good improvement over simple matching. 43 of the 47
queries in the best phrase matching run gave a precision of 100\% at 10\%
recall. Inspection of the remaining results shows that the errors could
typically only be fixed with a richer semantic representation allowing
interaction between the meaning of the words. For example, the query {\em
plastic toys} fails to match {\em plastic sword} because a sword is not
normally a toy. The precision at 5 documents shows less of an improvement as a
result of the small numbers of relevant captions.

Note that due to the lack of sources of good quality relevance judgements for this
kind of application, the results should be taken as suggestive of the
quality of phrase matching rather than as a definitive statement. An
evaluation was also carried out using the data from Smeaton and Quigley \cite{SQ:1996}, but we
concluded that the results could not be trusted, because the relevance
judgements were made against the images rather than the captions, and both the
captions and queries were of relatively low quality. In some cases we found
pairs of almost identical captions, one of which was judged relevant and one
irrelevant by Smeaton and Quigley's assessors. For comparison, the best
precision at 10\% recall reported by Smeaton and Quigley is around 62\%.

\section{Context extraction}

Context extraction is a means of obtaining additional information about
phrases which matched, by using the unmatched parts of the caption which are
close in the dependency structure to parts which did match. For example, if
the query was {\em camera lens}, and the captions included {\em long camera
lens} and {\em camera lens on a table}, then the contexts would be {\em long}
and {\em on a table}. Context extraction becomes valuable when there are many
retrieval results. Captions with similar contexts can be grouped together, for
example as shown the bottom half of in figure~\ref{c3}. A user can therefore
select or reject several retrieval results in one go by examining just the
contexts.

The algorithm for extracting the context is quite straightforward. It is
outlined in figure~\ref{c2}.
\begin{figure*} 
\centering
\begin{verbatim}
Query = 'camera with a lens.'
5 results:
SCORE  CAPTION
1      black SLR camera, with zoom lens, on a white surface.
       * camera: black, SLR, on a white surface
       * lens:   zoom
1      old-style black camera, with protruding lens, on a white surface.
       * camera: black, on a white surface
       * lens:   protruding
0.588  old camera, hip flask, box and album filled with sepia photographs.
       * camera: old
0.5    Canon camera, magnifying lens and fashion magazine on grey ridged surface.
       * camera: Canon, on a grey ridged surface
0.1    an astronaut floating within a space craft, showing the on-board cameras.
       * cameras: on-board
\end{verbatim}
\caption{Example ANVIL query result}
\label{fout}
\end{figure*}

\begin{figure*} 
\centering
\begin{verbatim}
let P be the set of path rules (input)
let T be the set of current words, initialised to all matched words (input)
let U be the set of available words, initialised to all unmatched words (input)
let S be the set of contexts, intially empty (output)

while T is not empty
{
    select a word t from T

    for each word u in U
    {
       if there is a context rule <rt,rv,rp,ru,rC> in P such that
           has_pos(t,rt)
           AND in_var(t,rv)
           AND has_pos(u,ru)
           AND on_path(t,u,rp)
       then
           find the smallest phrase C such that valid_phrase(C,rC,u)
           if there is such a C then
               add the context <t,C> to S
               remove u from U
    }

    remove t from T
}

where
  has_pos(t,rt)   if t has part of speech rt
  in_var(t,rv)    if t is stored in variable rv
  on_path(t,u,rp) if the path rp connects t and u
\end{verbatim}
\caption{The context extraction algorithm}
\label{c2}
\end{figure*}

The algorithm uses pre-defined context rules of the form $\langle
r_{t},r_{v},r_{p},r_{u},r_{C} \rangle$. In essence, it looks for words which
successfully matched, have a given part of speech $r_{t}$ and are stored in a
variable $r_{v}$. It then follows a path $r_{p}$ through the dependency structure,
arriving at an unmatched word $u$ with part of speech $r_{u}$, and then
extracts the syntactic context around it using the phrase type $r_{C}$ (for
example, {\em PP}, prepositional phrase). The restriction to the smallest
phrase is simply for cases where a phrase of a given type embedded within
another phrase of the same type. The elements of the rule can be wildcards,
which match anything. The algorithm delivers a set of pairs, each of a matched word
and its context. Simpler versions of the rules which do not have all of these
elements might also be possible.

An example path rule is $\langle noun,*,mod,*,* \rangle$, which
selects all modifiers of nouns. A rule of this sort might be used for
extracting {\em long} in the examples above. To get the context {\em on the
table}, a suitable rule might be $\langle noun,*,phead:prep,*,PP \rangle$,
i.e. from a matched word, follow a {\em prep} link followed by a {\em phead}
link, and select the {\em PP} surrounding the resulting word. Some example
contexts resulting from these rules are shown in figure~\ref{c3}, for the
query {\em camera with a lens}. The bottom half of the figure shows the
results gathered together by context. Presenting the results to the user in
this way would allow selection or rejection of several results with a single
decision, thus making it easier to manage large result sets.
\begin{figure} 
\centering
\begin{verbatim}
Query = camera with a lens

Captions and contexts
---------------------
  Camera with a lens
    {none}

  Large camera with a lens
    <camera [mod], large>

  camera with a lens on a table
    <camera [phead:prep], [on a table]PP>

  large camera with a zoom lens
    <camera [mod], large>
    <lens [mod], zoom>

  camera on a table with a long zoom lens
    <camera [phead:prep], [on a table]PP>
    <lens [mod], zoom>
    <lens [mod], long>

Captions gathered by context
----------------------------
  camera with a lens:
    camera modifiers:
      {none}     (1)
      on a table (2)
    lens modifiers:
      large      (1)
      zoom       (2)
      long       (1)
\end{verbatim}
\caption{Example contexts}
\label{c3}
\end{figure}

Perhaps the most important point about context extraction is not the algorithm
or exactly what the results look like, but the use of NLP to provide extract
information. Although there is some work in IR in extracting relevant parts of
the text, for example using named entity extraction, in general IR systems
just output a ranked list of matching documents. Context extraction
demonstrates that using NLP, which works with more detailed information
structures than traditional IR, we can produce a richer form of
output.

\section{Discussion}

The approach most closely related to phrase matching is that of
Sheridan and Smeaton \cite{SheSme:1992}. They start by constructing a
dependency tree (of a different form to ours), in
which interior nodes can be labelled, for example to mark the head or record
the preposition which links words on the nodes under it. The matching process
looks for pairs of words which are syntactically related in the query tree,
and which both appear in the tree for the key (caption). The nearest parent
nodes for the pairs of words are then checked for compatibility. Any parts of
the dependency structure which hang off the paths to the parent node, called
the residual structure, are examined to see if they could disrupt the
matching. For example, if words were both nouns, a verb in the residual would
block the match, since its presence indicates the nouns cannot stand in a
head/modifier relationship. The whole process is launched by looking at the
rightmost node in the query structure. A score is assigned based on the
proportion of words which match, possibly modified by certain residual nodes.

The main way in which this differs from our algorithm is that the selection
of nodes to try is {\em ad hoc}, rather than being guided directly by the
modification structure. The use of rules with a reduced score (such as {\tt
head = mod[]} above) and mopping up rules is also more explicit and modular
than the use of residuals. Furthermore, the scoring process in our phrase
matching takes the depth through the the structure (and hence the significance
of the terms) into account better, and is arguably more perspicuous. Some
further related work can be found in Schwarz \cite{Schw:1990}, in in which
syntactic structures are first converted to a normal form and then compared.

The work was conducted before the rise of interest in question-answering
(Voorhees and Tice \cite{Voor:1999}) which also uses short, precise queries to locate specific
information. Most of the TREC-8 question-answering systems used IR followed by entity
extraction, and one important limitation of this technique when applied to the
application described here is worth noting. The entity extracted as the answer
can appear anywhere in the retrieved text and consequently could part of some
modifying phrase rather than the main point of the caption, and so result in
retrieving images which do not correspond well to the request. By contrast,
the phrase matching rules can penalise such matches, provided the captions
model the content of the images well.

Two challenges follow. The first is to adapt techniques of this sort to full text
documents, in which there is a much richer linguistic structure, and where different
parts of the text may have different information content (a title compared to
a sentence in parentheses, for example). Secondly, there is a need to use
evaluation measures which place more emphasis on interactive retrieval and user reaction.
The assumption in much IR is that the results are simply
judged by their relevance to the user's information needs, essentially as a
binary decision. With an extension such as context extraction, where the
retrieval results contain extra information over the original data, we need an evaluation technique which is able to take into
account the benefit obtained from the results by the information user.

\section*{Acknowledgements}

The algorithms were implemented by the author and Aaron Kotcheff. The ideas
also benefitted from discussions with Tony Rose and Amanda Clare, and (at an
earlier stage) Tom Wachtel and Evelyn van de Veen.

\bibliographystyle{abbrv}
\bibliography{pm}

\begin{thebibliography}{10}

\bibitem{Elworthy:2000}
D.~Elworthy.
\newblock A finite state parser with dependency structure output.
\newblock In {\em 6th International Workshop on Parsing Technology, Trento,
  Italy}, 2000.

\bibitem{Fagan:1987}
J.~L. Fagan.
\newblock {\em Experiments in Automatic Phrase Indexing for Document Retrieval:
  A Comparison of Syntactic and Non-syntactic Methods}.
\newblock PhD thesis, Department of Computer Science, Cornell University, 1987.

\bibitem{Flank:1998}
S.~Flank.
\newblock A layered approach to {NLP}-based information retrieval.
\newblock In {\em Proceedings of 36th ACL and 17th COLING, Montreal}, pages
  397--403, 1998.

\bibitem{JiCon:1997}
J.~Y. Jiang and D.~W. Conrath.
\newblock Semantic similarity based on corpus statistics and lexical taxonomy.
\newblock In {\em Proceedings of ROCLING X, Taiwan}, 1997.

\bibitem{Jarvinen:1997}
T.~J{\aum}rvinen and P.~Tapanainen.
\newblock A dependency parser for {E}nglish.
\newblock Technical report TR-1, Department of General Linguistics, University
  of Helsinki, 1997.

\bibitem{pollock}
A.~Pollock and A.~Hockley.
\newblock What's wrong with internet searching.
\newblock {\em D-Lib Magazine}, March 1997.

\bibitem{Resnik:1995}
P.~Resnik.
\newblock Using information content to evaluate semantic similarity in a
  taxonomy.
\newblock In {\em Proceedings of the 14th International Joint Conference on
  Artificial Intelligence, Montreal}, 1995.

\bibitem{Rose:2000}
T.~Rose, D.~Elworthy, A.~Kotcheff, A.~Clare, and P.~Tsonis.
\newblock {ANVIL}: a system for the retrieval of captioned images using {NLP}
  techniques.
\newblock In {\em CIR2000: Third UK Conference on Image Retrieval, Brighton},
  2000.

\bibitem{Schw:1990}
C.~Schwarz.
\newblock Automatic syntactic analysis of free text.
\newblock {\em Journal of the Americal Society for Information Science},
  41(6):408--417, 1990.

\bibitem{SheSme:1992}
P.~Sheridan and A.~F. Smeaton.
\newblock The application of morpho-syntactic language processing to effective
  phrase matching.
\newblock {\em Information Processing and Management}, 28(3):349--369, 1992.

\bibitem{Sleator:1991}
D.~D.~K. Sleator and D.~Temperley.
\newblock Parsing {E}nglish with a link grammar.
\newblock Technical report CMU-CS-91-196, School of Computer Science, Carnegie
  Mellon University, 1991.

\bibitem{Smeaton:1997}
A.~F. Smeaton.
\newblock Information retrieval: Still butting heads with natural language
  processing?
\newblock In M.~T. Pazienza, editor, {\em Information Extraction: A
  Multidisciplinary Approach to an Emerging Information Technology}.
  Springer-Verlag, 1997.

\bibitem{SQ:1996}
A.~F. Smeaton and I.~Quigley.
\newblock Experiments on using semantic distances between words in image
  caption retrieval.
\newblock In {\em Proceedings of 19th SIGIR, Zurich}, pages 174--180, 1996.

\bibitem{Strz:1994}
T.~Strzalkowski.
\newblock Robust text processing in automated information retrieval.
\newblock In {\em Proceedings of the 4th Conference on Applied Natural Language
  Processing, Stuttgart}, pages 168--173, 1994.

\bibitem{Voor:1999}
E.~M. Voorhees and D.~M. Tice.
\newblock The {TREC}-8 question answering track evaluation.
\newblock In {\em Proceedings of the 8th Text Retrieval Conference}, 1999.

\end{thebibliography}

\end{document}